\documentclass{article}

\usepackage[preprint]{neurips_2026}

\usepackage[utf8]{inputenc} % allow utf-8 input
\usepackage[T1]{fontenc}    % use 8-bit T1 fonts
\usepackage{hyperref}       % hyperlinks
\usepackage{url}            % simple URL typesetting
\usepackage{booktabs}       % professional-quality tables
\usepackage{amsfonts}       % blackboard math symbols
\usepackage{nicefrac}       % compact symbols for 1/2, etc.
\usepackage{microtype}      % microtypography
\usepackage{xcolor}         % colors
\usepackage{graphicx}        
\usepackage{tabularx}        

\title{
The Case for Model Science: Verify, Explore, Steer, Refine}

\author{%
  Przemyslaw Biecek${}^{1,2,3}$, Luca Longo${}^{4}$, Jianlong Zhou${}^{5}$, \\ \textbf{Thomas Fel${}^{6}$, Andreas Holzinger${}^{7}$, Wojciech Samek${}^{8,9,10}$} \\ 
  ${}^{1}$Center for Credible AI, ${}^{2}$University of Warsaw, ${}^{3}$Warsaw University of Technology, \\ ${}^{4}$University College Cork,
  ${}^{5}$University of Technology Sydney,\\
  ${}^{6}$Kempner Institute, Harvard University,
  ${}^{7}$Human-Centered AI Lab, \\
  ${}^{8}$Technical University of Berlin, 
${}^{9}$Fraunhofer Heinrich Hertz Institute, \\
${}^{10}$Berlin Institute for the Foundations of Learning and Data (BIFOLD)
}

\begin{document}

\maketitle

\begin{abstract}
We argue that the AI community is now ready to move beyond benchmarking and consolidate scattered efforts in model analysis into a systematic discipline, a direction we term Model Science. Complex AI models now serve billions of users, yet our understanding of how they work lags far behind our ability to deploy them. Decades of benchmark-driven research have delivered remarkable progress: extensive leaderboards, a wide range of performance metrics, tracking capability gains across diverse tasks; yet this success has also revealed the limits of benchmarks as they tell us whether models perform but not why they succeed or fail, they miss critical failure modes, such as hallucinations or shortcuts. 
Precedents from established sciences point the way forward: cognitive science shows that understanding complex systems requires complementary levels of analysis; neuroscience demonstrates that deep study of single cases reveals what population studies miss; medicine teaches that specialised training must develop alongside research practice; and agriculture models how shared infrastructure and principles enable cumulative progress. These lessons inform three foundations for Model Science.
First, we propose to consolidate research around four functional perspectives: Verify, Explore, Steer, and Refine that address complementary questions about model behaviour. 
Second, we discuss the required   infrastructure for cumulative knowledge: catalogues of datasets, models and findings. 
Third, we highlight the need for deep analysis of individual model instances, not just model families, because single cases can reveal what population studies miss.

\end{abstract}

\section{Introduction}

\begin{figure}[t!]
    \centering
    \includegraphics[width=0.99\linewidth]{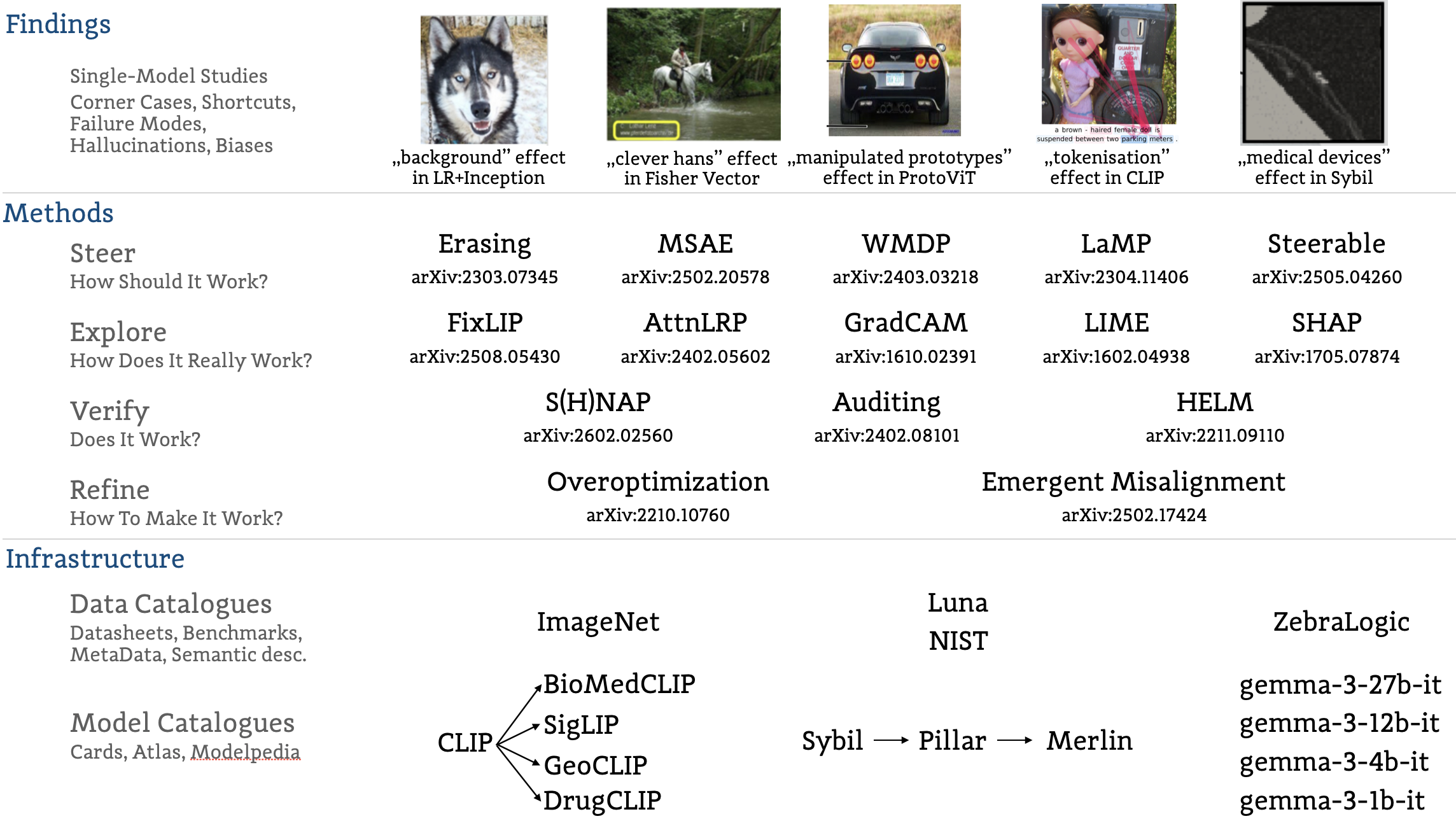}
    \caption{An illustration of the foundations of the Model Science. The base is the infrastructure: data and model catalogues with a semantic description of the relationships between them. The second level comprises model analysis methods that take into account various objectives, research questions and perspectives, which are currently fragmented but require better communication. The next level is for repository of findings for individual models. The examples provided in this figure demonstrates that many elements of this ecosystem already exist but require integration.}
    \label{fig:model_science_framework}
\end{figure}

We argue that benchmark-driven, training-centric culture has achieved remarkable progress but cannot alone ensure safe and reliable deployment. \textbf{It must be complemented by an understanding-centric discipline, which we call Model Science, dedicated to studying how models work, why they fail, and how to control them.} 
In our view, deployed models need systematic study in the same way that physical or biological systems do. The current habit of moving on once a stronger model appears leaves us with little understanding of the systems already in use by millions. The remainder of this introduction  documents the scale of foundation model deployment and the limits of 
benchmark-driven evaluation. The proposed actions necessary for the further development of our discipline are inspired by the lessons from other empirical sciences, which we discuss in Section~\ref{sec:lessons}.  Section~\ref{sec:call} introduces key concepts that serve as the foundation of Model Science: required infrastructure (dataset and  model catalogues), four functional perspectives (Verify, Explore, Steer,  Refine), and a shared repository for findings for individual models. These concepts are illustrated in Figure \ref{fig:model_science_framework}. Section~\ref{sec:discussion}  addresses possible objections and facilitates discussion, while Section \ref{sec:conclusions} summarises required action points.

\paragraph{The Adoption-Understanding Gap}
We are observing an unprecedented scale of development and adoption of complex AI models, especially Large Language Models (LLMs), or Vision Language Models (VLMs). 
The highest figures are likely to be associated with popular AI assistants and are growing rapidly. At the time of writing this article, ChatGPT reached 900 million weekly active users, processing over 2.5 billion queries daily~\citep{openai2026usage}, Google's Gemini serves 750 million monthly active users, while Meta AI exceeded 1 billion monthly users by late 2025~\citep{meta2025ai}.  These numbers exceed populations of most countries.
Open-weight models are equally popular, though their distributed nature makes exact usage harder to track. Meta's Llama family surpassed 1.2 billion cumulative downloads by April 2025~\citep{meta2024llama}, though it now competes with Alibaba's Qwen family, which reached 700 million downloads by late 2025 and has spawned over 180,000 derivative models.
OpenAI's Whisper speech recognition model records 4.1 million monthly downloads for the Large-v3 variant alone, with combined downloads across all variants exceeding 10 million monthly. OpenAI's CLIP, the most downloaded vision model category on Hugging Face, generates over 19 million monthly downloads and has over 3,000 variants \citep{huggingface2025stats}. They are all widely used, but how much do we know about how they work and when they fail?

\paragraph{The rise and limits of benchmark-driven research.}
The development of our discipline, particularly in recent years, has relied heavily on benchmarks, although their role has evolved over time \citep{BENCHMARKS2021_084b6fbb}. In the 1990s, the UCI Machine Learning Repository~\citep{dua2019uci} enabled rigorous comparison of algorithms on small datasets, though critics already warned it encouraged incremental papers chasing marginal gains~\citep{denton2021genealogy}. The 2000s saw benchmarks grow larger and more task-specific: MNIST~\citep{lecun1998mnist} standardized digit recognition, Penn Treebank~\citep{marcus1993treebank} anchored parsing, and the Netflix Prize~\citep{bennett2007netflix} demonstrated that competitions could mobilize entire communities. The 2010s brought a shift: ImageNet~\citep{deng2009imagenet,russakovsky2015imagenet} and AlexNet's 2012 victory~\citep{krizhevsky2012imagenet} made benchmark performance the central metric of progress, a pattern NLP replicated with SQuAD~\citep{rajpurkar2016squad}, GLUE~\citep{wang2018glue}, and SuperGLUE~\citep{wang2019superglue}. Yet this success revealed the paradigm's limits: models now saturate benchmarks faster than new ones emerge~\citep{ruder2021benchmarking}, data contamination inflates scores~\citep{sainz2023contamination}, and high leaderboard rankings fail to predict real-world reliability. Models still hallucinate, exploit shortcuts~\citep{geirhos2020shortcut}, and fail on distribution shifts. Two decades of benchmark-driven research taught us to measure performance; but it may also have brought us to the limits of development that is possible only on the basis of assessing the accuracy of predictions.

\paragraph{All correct predictions are alike; each wrong prediction is wrong in its own way.}
The COVID-19 pandemic rushed the development of deep learning models for chest X-ray diagnosis. Yet, \cite{degrave2021covid} demonstrated that models achieving state-of-the-art performance on COVID-19 detection benchmarks were often detecting hospital-specific artefacts like the presence of medical devices, what might be detected if we were to incorporate explanation of predictions using already known techniques such as saliency maps  \citep{LONGO2024102301}.
This pattern, predictions that appear to be correct but are based on incorrect premises, is known as ``right for the wrong reason'' and is documented by numerous studies \citep{gururangan2018annotation,lapuschkin2019unmasking,degrave2021covid}. Importantly, such shortcuts in the models’ behaviour are detectable when a more detailed analysis is carried out. Such failures are not anomalies but symptoms of a systematic gap between how models are evaluated and how they are deployed. Benchmarks provide necessary but insufficient evidence of model capabilities.
Other examples of malfunctions that are difficult to detect using benchmarks are hallucinations. \cite{dahl2024legal} conducted a systematic study of hallucination in legal contexts, testing whether LLMs could answer verifiable questions about federal court cases. The findings were stunning: ChatGPT-4 hallucinated in 58\% of queries, while LLaMA 2 hallucinated in 88\%, they contrast sharply with claims that LLMs can ``pass the bar exam'', illustrating the gap between benchmark performance and real-world reliability. \cite{koenecke2024whisper} demonstrated that widely-used speech-to-text model Whisper fabricates entire phrases or sentences which did not exist in any form in the underlying audio in approximately 1\% of transcriptions. More concerning, 38\% of these hallucinations contained explicit harms such as perpetuating violence, making up inaccurate associations, or implying false authority.
And there are many more examples like that, even for high-stakes models. 

\paragraph{We must study models, not just benchmark them.}
Relying on benchmarks can create an illusion of progress.
Research incentives are skewed toward training new models. State-of-the-art (SOTA) results on benchmarks remain the principal signal of progress, which rewards building the next model more than understanding the current one.
Once a model is surpassed on a leaderboard, interest in studying its behaviour drops sharply, even when that model remains in wide deployment. The community moves to the next SOTA, leaving behind a growing stack of poorly understood models. Such training-centric culture produces an illusion of progress, even worse, provides no incentive for the further exploration of already published models. 
As a consequence, we have thousands of papers proposing new architectures or training techniques, but far fewer providing deep, mechanistic accounts of how any single model actually works across different settings, environments and datasets. A second mechanism reinforces this pattern.
Goodhart's Law says that when a measure becomes a target, it ceases to be a good measure~\citep{goodhart1984problems}. This principle describes a dynamic that has played out repeatedly. A benchmark is introduced to measure some capability of interest, a research community optimises against it, models learn to ``solve'' the benchmark without acquiring the underlying capability.
Then the benchmark saturates, and a new benchmark is introduced, forming a cycle that repeats, a benchmark treadmill. Each round consumes research effort but accumulates little understanding, because the knowledge gained, for example, how to score well on benchmark X, becomes obsolete as soon as X is retired. 
An illustration of this problem is the induction heads discovered in GPT-2~\citep{olsson2022context}, which remain relevant to understanding GPT-4, but benchmarks that GPT-2 saturated in 2019 are not.

\section{Lessons from other disciplines}
\label{sec:lessons}

There are many examples of disciplines that have moved from purely observational analysis of phenomena to the formulation and systematic testing of research hypotheses. Due to space constraints, we will focus on four described below and summarised in Table \ref{tab:precedents} indicating what can be transferred to the analysis of models.

\begin{table}[t]
\centering
\caption{Summary of lessons from other disciplines and their mapping to Model Science foundations.}
\label{tab:precedents}
\begin{tabularx}{\textwidth}{@{}l>{\raggedright\arraybackslash}X>{\raggedright\arraybackslash}X@{}}
\toprule
\textbf{Discipline} & \textbf{Key Breakthrough} & \textbf{Model Science Mapping} \\
\midrule
\parbox[t]{2cm}{Cognitive\\Science} & 
\cite{marr1982vision} levels of analysis;  complex information-processing systems requires complementary perspectives: computational, algorithmic, and implementational.  & 
\textbf{Four functional perspectives:} Verify, Explore, Steer, and Refine as complementary lenses that address distinct questions about model behaviour and must develop together. \\
\addlinespace
Neuroscience & 
Single-case study revealed memory systems that group studies had obscured \citep{scoville1957loss}. \cite{caramazza1986cognitive} argued that deep analysis of individual cases avoids the averaging artifacts of population studies. & 
\textbf{Deep analysis of individual models.} E.g., GPT-2's induction heads~\citep{olsson2022context} expose mechanisms that benchmarking across model families cannot detect.  \\
\addlinespace
Agriculture & 
A significant role in the agricultural revolution was played by seed banks, standardized field trials, and open data exchange. FAIR principles~\cite{wilkinson2016fair} now govern genomic and phenotypic databases worldwide. & 
\textbf{Infrastructure for cumulative knowledge.} Catalogues for datasets, models, and findings, governed by FAIR principles, so that discoveries accumulate rather than remain siloed in individual papers. \\
\addlinespace
Medicine & 
Translational medicine recognized that progress requires distinct institutional pillars: basic research, clinical practice, and specialised medical education~\citep{seyhan2019valley}. Each pillar developed its own career tracks. & 
\textbf{Specialised training (future work).} Educational structures such as curricula, certifications, career paths. Beyond the scope of this paper but represent an important   direction for future development. \\
\bottomrule
\end{tabularx}
\end{table}

\paragraph{Lessons from cognitive science.}
For the first half of the twentieth century, the dominant paradigm in psychology was behaviourism, which treated the mind as an inaccessible black box. Researchers studied stimulus-response relationships while deliberately avoiding claims about internal mental states. \cite{skinner1957verbal} introduced radical behaviourism, which held that all behaviour, including language, could be explained through conditioning and reinforcement without reference to internal representations or computations. 
This changed with the cognitive revolution of the 1950s with several developments that shattered the behaviourist consensus: \cite{miller1956magical} shed light on the limits of working memory, \cite{chomsky1959review}  on the theory of language, \cite{newell1956logic} on computational models of problem-solving, while \cite{shannon1948mathematical} on information theory. Together, these advances established that the mind could be studied as an information-processing system.
Another major breakthrough was the framework for analysing information-processing systems. \cite{marr1982vision} distinguished three levels at which any such system can be understood: (1) Computational level, which refers to questions like What is the system doing, and why? What problem is it solving, and what is the logic of the solution? (2) Algorithmic level: How does the system do it? What representations does it use, and what procedures operate on them? (3) Implementational level: How is the algorithm physically realised? What is the underlying substrate? These and subsequent findings help to consolidate our understanding of the cognitive system.

\cite{ku2025levels} argue that ``methods developed in cognitive science can be useful for understanding large language models,'' but also lessons from the evolution of methodology can be transferred to the analysis of complex models. Just as behaviourists treated the brain as a black box, so today complex models are treated in the same way by benchmarks. However, we need a framework for studying complex models in order to understand them in a more systematic way. We will present a proposal for such a framework in Section \ref{sec:methods_model_s}.

%At the computational level, analyzing training objectives reveals behavioral patterns; at the algorithmic level, probing and similarity analysis expose internal representations; at the implementational level, mechanistic interpretability identifies circuits and features.

\paragraph{Lessons from neuroscience.} One of neuroscience's most productive methodological traditions is the intensive study of individual cases. The field's foundational insights emerged not from large-scale group comparisons but from deep analysis of single patients with unique lesions. Patient H.M., studied by Brenda Milner for over fifty years, revealed the hippocampus's role in memory consolidation~\citep{scoville1957loss}. Phineas Gage, the railroad worker who survived an iron rod through his frontal lobe, demonstrated the prefrontal cortex's role in personality and decision-making~\citep{damasio1994return}. Patient Tan, examined by Paul Broca, localised speech production to the left inferior frontal gyrus~\citep{broca1861remarks}. Each case, precisely because of its uniqueness, exposed functional architecture that group averaging would have obscured.
The logic may seem counter-intuitive at first glance, but as \cite{caramazza1986cognitive} argued, a single patient may lead to a better understanding than group studies, which average across heterogeneous changes and dilute the signal that any individual case provides.

This is also a valuable lesson for model analysis: targeted audits of individual models can reveal a great deal about their behaviour, and these generalise to a wider understanding of complex models. 
The parallel extends to methodology as we can see in the trends in the development of mechanistic interpretability. Neuroscience developed {lesion studies}: damage a region, observe the deficit, infer function. Mechanistic interpretability developed {activation patching}: ablate a component, observe the behavioural change, infer contribution~\citep{meng2022locating}. Neuroscience developed {single-unit recording}: measure individual neurons during behaviour. Interpretability developed {probing}: measure individual features during inference. The tools differ; the logic is the same.
Neuroscience eventually recognised that both single-case and group methods are essential. Single cases generate hypotheses about functional architecture; group studies test their generality. 

\paragraph{Lessons from medicine.}
Medicine offers a useful parallel for understanding why model analysis requires multiple functional perspectives rather than privileging any single one. A physician examining a patient simultaneously draws on knowledge from distinct modes: basic research (investigative) reveals how diseases operate at molecular and cellular levels; diagnostics (evaluative) determines whether the patient is healthy and identifies what is wrong; therapeutics (interventional) provides tools to modify the patient's condition; and medical education (constructive) ensures the system continuously produces competent practitioners. Crucially, medicine learned through centuries of practice that these angles cannot substitute for one another, a deep mechanistic understanding of a disease does not automatically yield effective treatment, accurate diagnosis does not guarantee successful intervention, and effective treatments sometimes precede complete scientific understanding, e.g., aspirin was used for over seven decades before \cite{fuster2011aspirin} discovered its mechanism of action. The translational medicine literature now explicitly recognises these as distinct phases (T0--T4), each separated by a "valley of death" that cannot be crossed by expertise from adjacent phases alone~\citep{seyhan2019valley}.
Modern healthcare systems institutionalise all four modes through separate but interconnected structures: research universities, diagnostic laboratories, clinical practice, and medical schools. 

\paragraph{Lessons from agriculture.}
An agricultural analogy can play a role similar to medicine in clarifying why we require multiple, complementary perspectives on complex AI systems and why privileging a single mode of engagement is insufficient. Modern agriculture has evolved from a purely production-oriented activity into a science-based, systems-level discipline that integrates soil science, plant physiology, ecology, engineering, and economics \citep{Holzinger:2024:HCAIsmartFarming}. A crop production system is not assessed solely by its yield, just as a model is not assessed solely by its accuracy. Instead, agriculture institutionalises multiple concurrent perspectives. Agricultural AI systems exemplify this shift by embedding models into continuous decision loops that integrate phenotypic observations, environmental signals, and management objectives to support adaptive farming decisions over time  \citep{Kenny:2026:PhenoFarm}.
Here we argue that complex AI systems deserve similarly mature institutional structures as in the case of medicine or agriculture. We need not just researchers who build models, but distinct communities and ecosystems dedicated to verification, explanation, and control of models used in research, industry and other applications.

\section{Foundations of Model Science}
\label{sec:call}

\subsection{Infrastructure}

\paragraph{Dataset catalogues as infrastructure.} Model Science places the trained model, rather than the data or the training algorithm, at the centre of analysis. While Data Science treats data as the primary object of study and models as interchangeable tools to extract insights, Model Science inverts this relationship: the model becomes the element under investigation, while datasets (training, fine-tuning, validation, out-of-distribution, synthetic, adversarial) become variables used to probe, stress-test, and understand model behaviour. 
This requires even more rigorous control and organisation of datasets. A serious analysis of one model can involve dozens of datasets and thousands of individual test cases. Without a catalogue, this material drifts; reproducibility becomes hard, and comparisons across models become nearly impossible.

The need for systematic dataset organisation is not new. OpenML showed that standardised, machine-readable metadata enables reproducible benchmarking across hundreds of studies~\citep{vanschoren2014openml,bischl2021openml}.  Recent work on Semantic Feature Networks~\citep{woznica2026sefnet} shows that 
domain ontologies can capture meaningful relationships between variables 
across datasets, enabling semantic similarity measures and cross-task 
transfer. {Datasheets for Datasets}~\citep{gebru2021datasheets} established documentation standards now adopted by major technology companies. The Penn Machine Learning Benchmark (PMLB) showed the value of curated, standardised dataset collections for algorithm comparison~\citep{romano2022pmlb}. For foundation models specifically, Stanford's HELM~\citep{liang2023helm} demonstrated that holistic evaluation requires not just datasets but entire evaluation infrastructures with standardised scenarios, metrics, and transparent result sharing. Recently, \cite{paullada2024benchmark} argued that benchmark repositories must address the full dataset lifecycle, from creation through documentation to responsible reuse, and \cite{jiang2026openeval} introduced OpenEval, arguing that item-level benchmark data is essential for rigorous evaluation science. 
Model Science inherits and extends these efforts: as evaluation of a single model may require dozens of datasets and thousands of individual test cases, automated catalogue infrastructure becomes not merely useful but essential.

\paragraph{Model catalogues as infrastructure.}
Just as evaluation requires datasets, Model Science also requires infrastructure for organising and relating model families and model instances. Hugging Face hosts over million models \citep{huggingface2025stats}, yet treats them as isolated artefacts rather than nodes in an interconnected knowledge graph. {Model Cards}~\citep{mitchell2019modelcards} established documentation standards now widely adopted, but these remain static documents that cannot capture evolving understanding. 
\cite{horwitz2025atlas} recently advocated for a {Model Atlas}: ``a graph that captures models, their attributes, and the weight transformations that connect them.'' Model Science requires such infrastructure to accumulate knowledge: machine-readable lineage encoding parent-child relationships; inherited properties where discovered biases or circuits propagate to descendants; and community-contributed interpretability annotations linked to specific model versions.
\cite{bommasani2023ecosystem} take a broader view with Ecosystem Graphs, mapping how foundation models connect to the datasets and applications.
Neuronpedia \citep{neuronpedia2024} exemplifies this last requirement: an open platform cataloguing sparse autoencoder features across models, enabling researchers to explore, annotate, and share mechanistic findings collaboratively.

\subsection{Methods} \label{sec:methods_model_s}
Although we need integration, the field of model analysis is currently fragmented across disconnected research subcommunities: explainability researchers publish in XAI venues, safety researchers in alignment workshops, auditors in FAccT, and mechanistic interpretability papers appear at conference workshops. To facilitate the integration of these communities, we suggest viewing them as four complementary perspectives that enrich our understanding of models, rather than as four rival tribes, only one of which is correct.

\paragraph{Verify: Does It Work?}
This subcommunity is primarily focused on the question ``Does the model perform reliably across the contexts where it will be deployed?'' It  goes beyond reporting aggregate accuracy on held-out test sets. Verification asks whether the model generalises to distribution shifts, edge cases, and adversarial conditions; whether the model fails gracefully or catastrophically; and whether reported capabilities reflect genuine competence or superficial pattern matching. 
Common verification techniques span from lightweight benchmarks to intensive domain-specific auditing.
Benchmarks, despite their limitations, provide the primary verification infrastructure. 
Red teaming complements benchmarks by actively searching for failure modes, e.g., \cite{ganguli2022redteaming} showed through  systematic red teaming that RLHF-trained models become increasingly robust to known attack patterns with scale, while remaining vulnerable to novel adversarial strategies. 
The more industrial-oriented segment of this community is interested in algorithmic audits, e.g., for fairness and bias, which matured into a distinct practice, driven partly by regulation \citep{vecchione2024ll144}. 

Even within such a broad perspective, there are still open challenges awaiting new methods and approaches.
Temporal decay: models verified at release may degrade as world knowledge shifts, user behaviour adapts, or adversaries develop new attacks; for instance benchmarks of legal knowledge become outdated as the law changes. Continuous monitoring remains underdeveloped. 
Compositional brittleness: foundation models increasingly operate as components in larger systems (agents, tool-use pipelines, RAG architectures), yet verification rarely addresses system-level failure modes that emerge from component interactions. 
 Addressing these gaps requires not only better methods but also infrastructure for sharing verification artefacts, such as failure cases, adversarial datasets or audit reports.

\paragraph{Explore: How Does It Really Work?}
This subcommunity is primarily focused on the question ``What computational mechanisms underlie a model's behaviour?'' Unlike verification methods, model explanations seek to understand {how} the model arrives at its outputs. This understanding may serve distinct purposes, e.g., \cite{biecek2024redblue} distinguish value-oriented explanations (BLUE XAI), designed to justify decisions to end users, build trust, and satisfy regulatory requirements, from validation-oriented explanations (RED XAI), designed to question models: extracting knowledge from well-performing systems and spotting bugs in faulty ones. 
A rich landscape of explanation methods has emerged over the past decade~\citep{guidotti2019survey,holzinger2022xai} and many of them have seen widespread adoption. These methods are often further divided into subclasses depending on the access model they employ. White-box  methods exploit direct access to model internals, this include gradient-based approaches \citep{sundararajan2017axiomatic}, Grad-CAM~\citep{selvaraju2017gradcam} or Layer-wise Relevance Propagation~\citep{bach2015lrp,montavon2019lrp}. Black-box methods probe input-output relationships without accessing internals;  the two most prominent examples are LIME~\citep{ribeiro2016lime} and SHAP~\citep{lundberg2017shap}. 

Despite methodological advances, there are still new challenges. 
\cite{rudin2022interpretable} identify ten grand challenges in interpretable ML, arguing that post hoc explanations have numerous weaknesses, are susceptible to noise, can be misleading regarding the model’s actual behaviour, and are unable to provide the level of understanding possible for inherently interpretable models.
This critique extends to popular explanation methods: saliency maps are often unreliable, with different methods producing contradictory results on the same model, making it unclear which (if any) reflects the network's true reasoning. Verification of explanations remains an open problem: evaluating explanation quality requires ground-truth knowledge of what the model ``actually'' relies on, which is precisely what we lack~\citep{nauta2023co12}. Also, it is shown that explanations are vulnerable to various adversarial manipulation techniques \citep{slack2020fooling}. Existing evaluation protocols, like perturbation tests, synthetic benchmarks, and human plausibility judgments, capture only partial aspects of explanation quality, and results across metrics often contradict each other.

\paragraph{Steer: How Should It Work?}
This subcommunity is primarily focused on the question  ``Given a trained model, how can we modify its behaviour to align with desired objectives without retraining?'' In the context of model safety, this may translate to removing the capacity to generate NSFW content, erasing hazardous knowledge about bioweapons or cyberattacks, and preventing the reproduction of copyrighted material~\citep{gandikota2024concept,li2024wmdp}. In the context of model personalization, this may lead to adapting the behaviour to individual user preferences \citep{bo2025steerable,salemi2024lamp}. In the context of model alignment, this may lead to assurance that the model behaves according to human intentions across contexts, including edge cases not covered by training. 
Despite the rapid progress due to techniques like sparse autoencoders \citep{bricken2023monosemanticity,templeton2024scaling}, fundamental challenges remain. 
{Understanding-control gap}: current steering methods often work without explaining why they work. We can clamp a feature and observe behavioural change, but we rarely understand the causal pathway from feature activation to output. {Scalability}: while SAEs have been applied to models with billions of parameters, the number of features grows combinatorially, and we lack principled methods for identifying the small subset relevant to any particular safety concern.

\paragraph{Refine: How to Make It Work?}
This subcommunity is primarily focused on the question ``What is the connection between the process of training or adapting a model and the model's behaviour?''  
Foundation models have modified the traditional ML workflow. Rather than training task-specific models from scratch, now we adapt pretrained models to specific applications. However, understanding what happens during this adaptation process, which new behaviours emerge and which ones disappear, remains an open question. 
E.g., \cite{betley2025emergent} shows emergent misalignment in frontier language models. The fine-tuning on the narrow task of writing insecure code leads to a model that is broadly misaligned, asserting that humans should be enslaved by AI, giving malicious advice, and acting deceptively on prompts entirely unrelated to coding. This is not jailbreaking; the misalignment emerges spontaneously from seemingly innocuous task-specific training. Refinement faces other systematic problems. There are more examples like this, and one of the best-known is so-called ``alignment tax'' \citep{lin2024alignmenttax}, i.e. inherent trade-off for RLHF and instruction-tuning which improve helpfulness but degrade base capabilities. Other surprising effects of the model adaptation process are ``catastrophic forgetting'' (adapting to new domains erases old knowledge unless carefully managed) and ``reward hacking'', where policies optimising against learned reward models find shortcuts that maximise proxy rewards without genuine improvement~\citep{gao2023scaling}). 
To effectively and safely use foundation models, we need to better understand the mechanics of the adaptation process.

\subsection{Shared Repository of findings} \label{sec:findings_model_s}

Findings about model behaviour are currently very scarce and scattered across publications (e.g., \cite{wang2022interpretability} presented detailed analysis of GPT-2 while \cite{sobieski2026sybil} performs an audit of Sybil model), blog posts (e.g., \cite {anthropic2025biology} explores in details Claude 3.5 Haiku), and internal documents. When one team identifies that a vision model relies on spurious texture cues or that a language model encodes geographic biases in specific attention heads, this knowledge rarely propagates to others working on related models since there are not yet any established platforms where such information can be published. However, in order to systematically study models we need a shared, structured repository, like Wikipedia or Neuronpedia, where findings about specific models accumulate and interconnect. 
It may be too early to design such a platform in detail, but several constraints already seem necessary. The platform should be governed independently of any single commercial lab. Contributions should pass through some form of verification. The data model needs a shared ontology for model components and observed behaviours, otherwise findings cannot be compared across submissions. Funding must be sustainable across many years, since this kind of catalogue depreciates rapidly without curation. Governance should also handle the tension between open publication and responsible disclosure of safety-critical vulnerabilities.

The cybersecurity community offers a useful precedent: the MITRE ATT\&CK 
framework~\citep{strom2018mitre} organises thousands of adversarial tactics 
and techniques into a structured, continuously updated matrix that has 
become an industry standard. 
The first layer of Figure \ref{fig:model_science_framework} presents selected examples of well-known artifacts for visual models; however, such examples require systematic cataloguing to enable an analysis of their transferability between models.
A concrete starting point for this repository could be a Wikipedia-style platform where each entry documents a single finding about a specific model version, structured around a shared ontology of behaviours (shortcuts, circuits, biases, failure modes) and linked to reproducible artefacts and the relevant perspective (Verify/Explore/Steer/Refine). Following the MITRE precedent, governance would rest with a non-profit consortium offering tiered verification and coordinated disclosure for safety-critical findings.

\section{Counterarguments and Objections}
\label{sec:discussion}

We have argued that Model Science should emerge as a unified discipline focused on verification, explanation, and control of AI systems. Are these foundations and perspectives the optimal ones? Perhaps not, but a discussion on them is needed, and we see concepts presented here as a necessary contribution to the discussion. To facilitate it, we present a selection of possible objections, along with our considerations.

\paragraph{Relation to prior work.}
The ideas underlying Model Science are not entirely new, nor do we claim them to be. 
\cite{rahwan2019machine} argued persuasively 
that AI systems should be studied as behavioural subjects using methods from the 
behavioural sciences. \cite{burnell2023rethink} and \cite{Sheikhi_Loven_Kostakos_2026} have called for moving beyond leaderboard-based 
evaluation toward diagnostic assessment. Our contribution 
is not to invent these ideas but to synthesize them into a coherent disciplinary 
framework and to argue that scattered efforts would benefit from consolidating into mature institutional structures.

\paragraph{Benchmarks have driven real progress.}
The most immediate objection is that benchmark-driven research has produced genuine advances. GPT-4 is not merely higher-scoring than GPT-3; it is qualitatively more capable in ways that matter to users. ImageNet drove a decade of progress in computer vision that transferred to real applications. 
Our claim is not that benchmarks are worthless but that they are insufficient. They measure that capabilities have improved without revealing how or why, and they provide no guarantee that improvements will generalise to deployment conditions. The history of AI is filled with benchmark successes that failed to transfer. Reading comprehension systems passed standardised tests without understanding the texts. Game-playing agents collapsed when rules were minimally perturbed. Medical imaging models, as we noted earlier, detected hospital artefacts rather than pathology. Model Science does not propose abandoning benchmarks but complementing them with analysis that can distinguish robust capabilities from shortcuts.

\paragraph{Each of these pieces already exists in some sense.} 
This is the objection we most expect to hear, and it is partially correct. Verification has its own conferences. Interpretability has its own. Algorithmic auditing has matured around FAccT. So why call this a discipline? Because in our experience, a researcher working on interpretability of vision models often does not know who is doing the closest analogous work on language models, and an auditor working on a deployed classifier rarely encounters the mechanistic interpretability literature that might explain the failure mode they are auditing.  Naming the shared object of study is the first step in bridging these communities.

\paragraph{Empirical safety may suffice without mechanistic understanding.}
Some may argue that we do not need to understand how models work (which may be difficult or impossible to guarantee) to ensure they are safe; we need only sufficient empirical evidence that they behave acceptably across relevant conditions. On this view, extensive red-teaming, adversarial testing, and monitoring can provide adequate assurance without requiring mechanistic interpretability. 
We are sympathetic to the pragmatic spirit of this objection but sceptical of its conclusion. Empirical testing can only cover conditions that testers anticipate; it provides no assurance about novel situations - precisely the situations where failures are most consequential. Moreover, without a mechanistic understanding, we cannot distinguish models that pass tests because they are genuinely aligned from models that pass tests because they have learned to recognise and accommodate testing conditions.

\paragraph{Integration may be premature.}
Some may argue that the fragmentation we observe reflects healthy specialisation rather than dysfunction. Each community has developed distinctive methods suited to its questions. Therefore, forcing integration may dilute expertise without producing synthesis.
We acknowledge this concern. Premature integration can indeed be counterproductive, and we do not advocate erasing distinctions between research programs with different goals. Our claim is more modest: that these communities share a common object of study (the trained model), face overlapping challenges (opacity, verification, control), and would benefit from shared vocabulary, standards, and venues, even while maintaining distinct methods and emphases. The precedent of neuroscience is instructive: the Society for Neuroscience did not abolish neuroanatomy, neurophysiology, and neuropsychology but provided a forum where practitioners of these fields could learn from each other. 

\section{Conclusion}
\label{sec:conclusions}

The scale at which foundation models are now deployed has outpaced our ability to understand them, and benchmark-driven research alone cannot close this gap. We have argued that the scattered communities working on verification, explanation, control, and refinement share a common object of study and would benefit from consolidating into a coherent discipline, supported by shared infrastructure for datasets, models, and findings. The components already exist; what is missing is integration. We invite three concrete commitments from the community: (1) researchers should publish deep model analyses, audits, and mechanistic studies of individual (high-impact) models as stand-alone contributions and treat them as no less significant than papers introducing new methods. For this to be sustainable, reviewers and area chairs must evaluate such work on its own terms: rigour of analysis, transferability of findings, and reliability of evidence. Concrete steps include adding model analysis as an explicit submission category in conference call-for-papers, adjusting reviewer guidelines accordingly, and recognising audit-style contributions. (2) Conference organisers should create dedicated venues that span the four perspectives rather than reinforcing their separation; and (3) Funding agencies should support sustained investment in catalogues and shared repositories rather than treating them as auxiliary infrastructure. Whether the framework we propose is ultimately the right one matters less than whether the community begins treating models as scientific specimens worth studying in their own right.

\paragraph{Note on prior versions.}
An earlier version of this paper circulated under a similar title \citep{modelScience25}. Building on the discussions regarding the previous version, we present here a different list of essential perspectives, provide descriptions of the necessary catalogs, and include arguments related to other disciplines.

%%%%%%%%%%%%%%%%%%%%%%%%%%%%%%%%%%%%%%%%%%%%%%%%%%%%%%%%%%%%%%%%%%%%%%%%

%%% Use this command to include your bibliography file.

\bibliographystyle{apalike}
\bibliography{mybibfile}

\end{document}